\RequirePackage{amsmath}
\documentclass{llncs}

\usepackage[utf8]{inputenc}
\usepackage[T1]{fontenc}
\usepackage{latexsym}
\usepackage{amssymb}
\usepackage{lmodern}
\usepackage{fix-cm}
\usepackage{url}
\usepackage{array}
\usepackage{geometry}
\usepackage{graphicx}
\usepackage{tabularx}
\usepackage{array}
\usepackage{mdwmath}
\usepackage{mdwtab}
\usepackage{float}
\usepackage{tikz}
\usetikzlibrary{shapes,arrows}
\usepackage{makecell}
\usepackage{diagbox}
\usepackage{eurosym}
\usepackage{eulervm}
\usepackage{palatino}
\usepackage{booktabs}
\newcommand{\pc}{\mathbf{P}}

\title{General Rough Modeling of Cluster Analysis}
\author{\textsf{A. Mani}}
\institute{Indian Statistical Institute, Kolkata\\
203, B. T. Road. Kolkata-700108, India\\
Email: \texttt{$a.mani.cms@gmail.com$, } \texttt{$amani.rough@isical.ac.in$}\\
Homepage: \url{https://www.logicamani.in}\\
Orcid: \url{https://orcid.org/0000-0002-0880-1035} }

\begin{document}

\maketitle

\begin{abstract}
In this research a general theoretical framework for clustering is proposed over specific partial algebraic systems by the present author. Her theory helps in isolating minimal assumptions necessary for different concepts of clustering information in any form to be realized in a situation (and therefore in a semantics). \emph{It is well-known that of the limited number of proofs in the theory of hard and soft clustering that are known to exist, most involve statistical assumptions}. Many methods seem to work because they seem to work in specific empirical practice. A new general rough method of analyzing clusterings is invented, and this opens the subject to clearer conceptions and contamination-free theoretical proofs. Numeric ideas of validation are also proposed to be replaced by those based on general rough approximation. The essential approach is explained in brief and supported by an example. 
\end{abstract}


\section{Introduction}

Hard and soft clustering processes are based on ideas of optimization that contribute to uncertainty, vagueness and indeterminacy in associated proofs and measures. Often convergence of algorithms or cluster validity are not proven, and the ones proved proceed from questionable statistical and topological assumptions \cite{bcmr2019,hmmr2016} about the context associated with a dataset. To see this consider the problem of clustering six objects into two clusters. Among other things, certain possible clusters may not be reasonable. In this situation, what does it mean to consider the collection of all possible clusters relative to a purely combinatorial perspective? This is just one of the ways in which statistical proofs may lose universality and relevance. Aggregation operations are also known to become paradoxical in the context of statistical tests and decision theory \cite{hd2003}. The level of context dependency in the use of clustering techniques in the AI (and ML) literature is severe -- an important heuristic is to try every technique that is known to work in related application contexts (or use cases). This scenario suggests that it can be useful to build a minimalist framework for exploring proofs, ontology, and associated methodology (and also because application contexts are loaded with excess baggage). 

The basic problem is of developing a reasonable framework with its assumptions and not of a language of expression (though the former requires some of the latter). Metric logics and variants (see for example \cite{dis2020}) in particular, cannot add much to the ideas of validation for distance-based clustering because they merely intend to express the same facts and methods in a restricted language. Higher order rough frameworks for analyzing soft and hard clustering are proposed in this research over axiomatic granular rough sets as they are far more capable of handling knowledge evolution. New concepts of clustering (including a definition) are also proposed.

The essence of the introduced frameworks revolve around the following ideas:
\begin{description}
\item[A]{A language (or model) that can express clustering related information should include at least one ternary predicate $\delta$ (with $\delta abc$ interpreted as $a$ is closer to $b$ than to $c$ in some sense).}
\item[B]{Since all kinds of clustering involve approximations of some kind (that may possibly be ontologically justified) in their definition, computation or validation, it is necessary to permit generalized approximations that are associated with the intrinsic structure of the data.}
\item[C] {It is important to avoid making wild external numeric approximations about the data, and contaminating it \cite{am5586,am501}. This is also about the data being able to speak by itself.}
\item[D]{Granular approximations (in the axiomatic sense) are better suited to handle meaning and evolution of knowledge (but a number of rough approximations may get excluded by restricting to higher granular operator spaces/partial algebras). }
\item[E]{Concepts of aggregation, and commonality can be partial (as they are in real life). It does not always make sense to combine objects or concepts for example. Further they may be unrelated to fundamental part-of relations in the context.}
\item[F]{While numeric valuations are best avoided, in some contexts they can be meaningful. Frameworks should be able to handle this possibility.}
\end{description}

Points \textbf{D, E, F} are already considered in previous work of the present author \cite{am5586,am501} in the context of higher granular operator spaces/partial algebras and variants. These will be used with minimal explanation.

\subsection{Background}

The reader is expected to be familiar with the literature on abstract granular approaches to general rough sets \cite{am5586,am501}, and some mereology \cite{am5586,ham2017,rgac15}. Further concepts such as those of higher granular operator spaces/partial algebras, contamination and admissible granulation will be assumed.  

The process or concept of \emph{cluster validation} generally refers to exploring the quality of one or more clustering methods and possibly comparing them. In almost all cases, true class information is not available (that is if one avoids looking at anything apart from the dataset) and validation methods are inherently not rigorous even in comparison to statistical ones used in supervised learning. Further they are subjective, highly contextual, are not generalizable, and  \emph{assume some heuristics that are not well understood and in some cases even the values produced may not be clear (see \cite{kr2005} for example)}. 

\emph{Rough clustering} refers to clustering methods that are enhanced with rough set theoretical ideas of what a rough cluster ought to be. These may sometimes be seen as a two layered process in which the last layer is about interpreting membership from a rough perspective. A rough cluster is seen as a pair consisting of a lower and upper approximation of an object or as some other representation of a rough object \cite{ivo2016,petersg2014,sm2004,zhoupm2011}. The present research  though related, is about building a general rough framework for all clustering contexts, and differs in purpose and methods.

\section{New Rough Semantic Approaches}

Let $S = \left\langle\underline{S},\, \Sigma^{\underline{S}} \right\rangle$ be a partial algebraic system over a signature $\Sigma$ (the superscript indicates interpretation on $\underline{S}$). For any subset $\Sigma_o$ of $\Sigma$, $S_o = \left\langle\underline{S},\, \Sigma_o^{\underline{S}} \right\rangle$ will be referred to as a \emph{reduct} of $S$. For two terms $s,\,t$, $s\,\stackrel{\omega}{=}\,t$ shall mean, if both sides are defined then the two terms are equal (the quantification is implicit), while $s\,\stackrel{\omega ^*}{=}\,t$ shall mean if either side is defined, then the other is and the two sides are equal (the quantification is implicit). \emph{As points can be regarded as singletons, the point-subset distinction that is explicitly assumed in clustering theory can be discarded}. That is the partial algebraic system can include all. Apart from parthood, ternary predicates of the form $\delta$ that satisfy some of the inner (i-coh) and near (n-coh) coherence conditions below are of interest:
\begin{align}
(\forall a, b)\, \delta bba \tag{i-coh}\\
(\forall a, b,c )\, (\delta abc \longrightarrow \delta bac ) \tag{n-coh}\\
(\forall a, b)\, \neg \delta abb \tag{i-coh-2}\\
(\forall a, b,c )\, (\delta abc \longrightarrow \neg \delta acb ) \tag{strict n-coh}\\
(\forall a, b,c )\, (\delta abc\,\&\, \delta aeb \longrightarrow \neg \delta aec ) \tag{trans-1}
\end{align}
Intended meanings of $\delta abc$ are \emph{$a$ is closer to $b$ than $c$ in some sense}, \emph{$a$ is more similar to $b$ than $c$ in some sense} and variants thereof. This predicate covers the intent of using metrics, similarities, dissimilarities, proximities, descriptive proximities, kernels and other functions for the purpose. 

For an arbitrary subset $K$ of a set $H$ to qualify as a cluster (under a large number of additional constraints), many conditions that depend on $\delta$ (essentially) are typically required to be satisfied. These can be written with restricted quantification as in 
$(\forall a, b\in K)(\forall c\in K^c) \, \delta abc$. But such an expression is neither elegant nor general enough or useful from a logical perspective. A better strategy is to identify clusters with a unary predicate $ \kappa$ in $H$ and express such conditions with an additional binary \emph{part of} predicate $\pc$. 

In describing clustering contexts or processes, it may not always be reasonable to combine objects (or groups formed at some step) into a new object, or their combination may be regarded as a plural object (like some collection of subsets of a union \cite{am5586}) and it can also happen that parthood does not define the mereological sum always \cite{ham2017}. Such a partial sum operation $\oplus$ can be expected to satisfy all of the following properties:
\begin{align}
a\oplus b \stackrel{\omega^*}{=} b\oplus a; \: a\oplus a \stackrel{\omega}{=} a;\: a\oplus (b \oplus c) \stackrel{\omega}{=} (a\oplus b)\oplus c    \tag{$\omega^*$-com; $\omega$-id; $\omega$-asso}\\
\delta abc \longrightarrow \delta (a\oplus a)bc; \:\delta abc \longrightarrow \delta a(b\oplus b)c;\: \delta abc \longrightarrow \delta ab(c\oplus c)   \tag{$\delta$-sum1; $\delta$-sum2; $\delta$-sum3}
\end{align}

In general clustering contexts, it is often the case that an additional external ordered algebraic system is used to measure (or evaluate) ideas of nearness or proximities (in a descriptive, spatial or generalized metric sense \cite{cgjp2016}). These may be partially ordered sets or even the semi-ring of positive reals. A minimalist structure that is always present (as an algebraic reduct) is a parthood space \cite{am5550} in which it is possible to express some idea of comparison (in a perspective of containment) that may not necessarily be transitive. 

The definition of $\delta$ can be made explicit with the help of additional maps $f: S^2 \longmapsto H$ (this may not be required as shown in the example below \ref{rme}) that satisfies one or more of
\begin{equation*}
\delta abc \longrightarrow \pc f(a, b) f(a, c); \: \pc f(a, b) f(a, c)  \longrightarrow \delta abc ; \: \delta abc \leftrightarrow \pc f(a, b) f(a, c) \tag{def1; def2; def0} 
\end{equation*}
In some cases as in clustering contexts that depend on a metric, the stronger version \textsf{def0} actually holds.

Given the above concepts, it is possible to specify a number of formula using higher order constructs that correspond to definitions of clusters. A specific general form is (with $A^*$ being a subobject of $A$ and $A^{\circ}$ being a generalized complement)
\begin{equation}
(\forall a\in A)(\forall b\in A^*)(\forall c\in A^\circ)\, \delta abc \tag{clue} 
\end{equation}
It is not the most general form because the dependence on specific clusters like $A$ takes other forms. Under additional conditions, the following form has the potential to specify clusters.
\begin{equation}
(\forall a\in A)(\forall b\in B)(\forall c\in E)\, \delta abc \tag{gclue} 
\end{equation}
$B$ in particular can depend on $A$, $E$ and possible ideas of being an $A$. The dependence between $B$, $E$ and $A$ might also be of a higher order nature in that the concept of a cluster is in relation to the collection of all clusters. Further additional properties may be satisfied by the collection of all clusters. The former can all be expressed with the help of an additional unary operation $\kappa: \, H \longmapsto H$ that enables identification of cluster members. 

Based on the above considerations, the following partial algebraic systems appears to be optimal for theoretical studies on clustering in the mentioned perspective.
\begin{definition}\label{mss}
A partial algebraic system of the form  $S = \left\langle\underline{S},\, \Sigma^{\underline{S}} \right\rangle$ with 
$\Sigma = \{\pc, \delta, \oplus, \kappa, \leq, \vee, \wedge, l, u, \top , \bot\}$ of type $(2,3,2,1,2,2, 2,1, 1, 0, 0)$
$l, u$ being operators $:\underline{S}\longmapsto \underline{S}$ satisfying the following ($\underline{S}$ is replaced with $S$ if clear from the context. $\vee$ and $\wedge$ are idempotent partial operations, $\kappa$ a unary predicate for identifying clusters, and $\pc$ is a binary parthood predicate) will be referred to as a \emph{minimal soft clustering system} (MSS) whenever the conditions \textsf{i-coh, n-coh, i-coh-2,trans1, clos1} and \textsf{PT1, PT2, G1, G2, G3, G4, G5, UL1, UL2, UL3, TB} of the definition of high granular operator space (GGS) \cite{am5586} hold:

\begin{align*}
(\forall x) \pc xx ; \: (\forall x, b) (\pc xb \, \&\, \pc bx \longrightarrow x = b) \tag{PT1; PT2}
\\
a\vee b \stackrel{\omega}{=} b\vee a ;\: a\wedge b \stackrel{\omega}{=} b\wedge a ;\:
(a\vee b) \wedge a \stackrel{\omega}{=} a ;\: (a\wedge b) \vee a \stackrel{\omega}{=} a \tag{G1; G2}\\
(a\wedge b) \vee c \stackrel{\omega}{=} (a\vee c) \wedge (b\vee c);\: 
(a\vee b) \wedge c \stackrel{\omega}{=} (a\wedge c) \vee  (b\wedge c) \tag{G3; G4}\\
(a\leq b \leftrightarrow a\vee b = b \,\leftrightarrow\, a\wedge b = a  ) \tag{G5}\\
(\forall a \in \mathbb{S})\,  \pc a^l  a\,\&\,a^{ll}\, =\,a^l \,\&\, \pc a^{u}  a^{uu} ;\: 
(\forall a, b \in \mathbb{S}) (\pc a b \longrightarrow \pc a^l b^l \,\&\,\pc a^u  b^u) \tag{UL1; UL2}\\
\bot^l\, =\, \bot \,\&\, \bot^u\, =\, \bot \,\&\, \pc \top^{l} \top \,\&\,  \pc \top^{u} \top ;\: 
(\forall a \in \mathbb{S})\, \pc \bot a \,\&\, \pc a \top    \tag{UL3; TB}
\end{align*}
\end{definition}

In the context of the above definition, if the condition \textsf{strict n-coh} (\textsf{lclu}) is also satisfied, then the MSS will be referred to as \emph{strict} (\emph{rough}).  
\begin{equation*}
(\forall a, b,c )\, (\delta abc \longrightarrow \neg \delta acb );\:(\forall a)(\kappa a \longrightarrow \kappa a^l) \tag{strict n-coh; lclu}\label{lclu}  
\end{equation*}
If the signature in Def \ref{mss} is $\Sigma^* = \{\pc, \gamma, \delta, \oplus, \kappa, \leq, \vee, \wedge, l, u, \top , \bot\}$ instead and the additional conditions on the granulation of a GGS \cite{am5586} are also satisfied, then the resulting system will be referred to as a \emph{granular MSS} (GMSS). Granularity in this sense is essential for construction of knowledge as in \cite{am5586}.

The conditions defining admissible granulations mean that every approximation is representable by granules in an algebraic way, that every granule coincides with its lower approximation (granules are lower definite), and that all pairs of distinct granules are part of definite objects (those that coincide with their own lower and upper approximations).

Internal measures that characterize the quality of clusters in terms of deviance from associated approximations (and therefore lack of coherence) in a \textsf{GMSS} $\mathbb{H}$ in which $\vee$, and $\wedge$ are set union and intersection respectively and $\setminus$ is a partial set difference operation are defined below:
\begin{definition}\label{setclu}
In the GMSS $\mathbb{H}$ mentioned, let $\mathcal{C}$ be a clustering on $\top$. then the \emph{lower deficit} ($C^{\flat}$) of a cluster $C\in \mathcal{C}$ will be the set $(C\setminus C^l)^u$ (if defined), and its \emph{upper deficit} ($C^{\eth}$) will be the set $(C^u\setminus C)^u$ (if defined). Further $C$ will be \emph{lu-valid} iff $c^l = c^u = C$, \emph{l-pre-valid} if and only if $(\exists V\in \mathbb{S}) V^l = C$, and \emph{l-traceable} if and only if $(\exists V\in \mathbb{S}) V = C^l$. Analogous concepts of $u$-pre validity can be defined. 
In addition, if all clusters in $\mathcal{C}$ are $l$-pre-valid (resp. $lu$-valid, $u$-pre-valid, $l$-traceable, $u$-traceable) then $\mathcal{C}$ will itself be said to be $l$-\emph{pre-valid} (resp. $lu$-valid, $u$-pre-valid, $l$-traceable, $u$-traceable). 
\end{definition}
\begin{proposition}
In the context of Definition \ref{setclu}, if the $l$-deficit (resp. $u$-deficit) of a cluster $C$ is computable, then it must necessarily be $l$-traceable (resp. $u$-traceable). 
\end{proposition}
The central idea of $lu$-validity (and weakenings thereof) is that of representability in terms of granules and approximations. These do not test the key predicate $\delta$ for validation, and the aspect is left to the process of construction of rough approximations. By contrast, the $*$-deficits are an internal measure of what is lacking or what is in excess. 

\section*{Delta Methodology} 
\emph{The framework of \textsf{MSS} introduced permits evaluation of clusterings relative to parthood and approximation operations. Because the predicate (or equivalent partial operation) and approximations are constructed by the user, the system is not inherently constraining in any way}. The proposed \emph{delta methodology} consists of the following steps:
\begin{description}
\item [Step-1]{Define most of the MSS for the context (except for $\kappa$ and $\delta$ possibly). In addition associate mereo-ontologies with the system.} 
\item [Step-2]{Do feature selection if required and form a reduct of the original MSS}
\item [Step-3]{Compute clusters as per desired algorithm}
\item [Step-4]{Either define the new clustering in the MSS or form a new MSS with the clustering}
\item [Step-5]{Investigate through minimal additional assumptions and possible definitions of $\delta$.}
\end{description}

\textbf{Delta Methodology: Example}\label{rme}

For this example, the reader needs to refer to section 6.3.3 of \cite{am501} by the present author. Let $H \, =\,\{x_{1}, x_{2}, x_{3}, x_{4}\}$ and $T$ a tolerance on it generated by $\{(x_{1}, x_{2}),\,(x_{2},x_{3})\}$. Denoting the statement that the granule generated by $x_{1}$ is $(x_{1},\,x_{2})$ by $(x_{1}:x_{2})$, let the granules be the set of predecessor neighborhoods:  $\mathcal{G}=\{(x_{1}:x_{2}),\,(x_{2}:x_{1},x_{3}),\,(x_{3}:x_{2}),\,(x_{4}:)\}$. The different approximations (lower ($l$), upper ($u$) and bited upper ($u_b$)) are then as in \cite{am501} (the symbols are changed here, and $\sim$ is an equivalence on $\wp(H)$ defined by $A\,\sim\,B \text{ if and only if } A^l \, =\,B^l \& A^{u_b} \, =\,B^{u_b}$. Now let, $\underline{S} = \wp(H)$, $\pc = \subseteq = \leq$, and consider the following possible definitions of $\delta$:
\begin{align*}
\delta abc \text{ if and only if } \pc (a\cup b)(a\cup c)\, \&\, \neg \pc (a\cup c) (a\cup b)   \tag{E1}\\
\delta abc \text{ if and only if } \pc (a\cap c)^l(a\cap b)^l\, \&\, \neg \pc (a\cap b)^l (a\cap c)^l    \tag{E2}\\
\delta abc \text{ if and only if } \pc (a\cup b)^u(a\cup c)^u \tag{uE1}\\
\delta abc \text{ if and only if } \pc (a\cup b)(a\cup c)   \tag{E0}
\end{align*}
Suppose some clustering technique produces the clustering $\mathcal{C}$ defined by:\\ $\mathcal{C} = \{\{x_1, x_3\},  \{x_2, x_3\}, \{x_2, x_4\}\}$. \emph{Then one can see that it is compatible with definitions E1 and E0}. Further the \emph{lower and upper deficits} of the $\{x_2, x_4\}$ are $\{x_1, x_2, x_3\}$ and $\{x_1, x_2, x_3\}$ respectively. Note that the granulation $\mathcal{G}$ can also be seen as a clustering relative to uE1. In the context, the MSS is $S= \left\langle \underline{S}, \subseteq, \delta, \oplus, \kappa, \cup, \cap, l, u, H , \emptyset \right\rangle$, with $\kappa$ being specified by $\mathcal{C}$. It is also easy to extend it to a GMSS.

\textbf{Future Work:} A much extended version of this research for a logic-oriented audience is under revision. This includes a detailed critique of the issues with existing cluster validation techniques. Newer methods in the context of this research based on prototypes, rationality and other criteria and novel rough clustering methods are part of forthcoming papers of the present author. A joint paper on related empirical studies in education research is in progress. Researchers and practitioners take a soft view of validation in both soft and hard clustering. The proposed framework affords a better way of formalizing the soft aspect. More work on this is obviously motivated. 
\begin{small}
\begin{flushleft}
\textbf{Acknowledgment:} This research is supported by a Women Scientist grant of the Department of Science and Technology.
\end{flushleft}
\end{small}
\bibliographystyle{splncs.bst}
\bibliography{algroughf69fl}

\end{document}